\documentclass[10pt,twocolumn,letterpaper]{article}

\usepackage{wacv}
\usepackage{times}
\usepackage{epsfig}
\usepackage{graphicx}
\usepackage{amsmath}
\usepackage{amssymb}

\usepackage{makecell}

\def\wacvPaperID{****} % Enter the WACV Paper ID here

\wacvfinalcopy % *** Uncomment this line for the final submission

\ifwacvfinal
\def\assignedStartPage{1} % *** Enter the assigned starting page number (instead of 9876)
\fi

\ifwacvfinal
\usepackage[breaklinks=true,bookmarks=false]{hyperref}
\else
\usepackage[pagebackref=true,breaklinks=true,colorlinks,bookmarks=false]{hyperref}
\fi

\ifwacvfinal
\else
\fi

\begin{document}

%%%%%%%%% TITLE
\title{On Evaluating Weakly Supervised Action Segmentation Methods}

\author{Yaser Souri$^\dag$, Alexander Richard$^\dag$, Luca Minciullo$^\S$, Juergen Gall$^\dag$\\
$^\dag$ University of Bonn, $^\S$ Toyota Motor Europe\\
{\tt\small \{souri, richard, gall\}@iai.uni-bonn.de, luca.minciullo@toyota-europe.com}
}

\maketitle
%\thispagestyle{empty}

%%%%%%%%% ABSTRACT
\begin{abstract}
   Action segmentation is the task of temporally segmenting every frame of an untrimmed video. Weakly supervised approaches to action segmentation, especially from transcripts have been of considerable interest to the computer vision community. In this work, we focus on two aspects of the use and evaluation of weakly supervised action segmentation approaches that are often overlooked: the performance variance over multiple training runs and the impact of selecting feature extractors for this task.
   To tackle the first problem, we train each method on the Breakfast dataset 5 times and provide average and standard deviation of the results. Our experiments show that the standard deviation over these repetitions is between 1 and 2.51\%, and significantly affects the comparison between different approaches.
   Furthermore, our investigation on feature extraction shows that, for the studied weakly-supervised action segmentation methods, higher-level I3D features perform worse than classical IDT features.
\end{abstract}

%%%%%%%%% BODY TEXT
\section{Introduction}
Action segmentation is the task of temporally segmenting every frame of an untrimmed video. It is usually studied in the context of instructional activities performed by a single person \cite{breakfast,50salads,crosstask} and fully supervised approaches for action segmentation achieve already very good results \cite{kuehne2016end,lea2016segmental,lea2017temporal,MS-TCN}.
As providing frame-level supervision for action segmentation is an expensive task, weakly supervised approaches have received a lot of attention from the community \cite{hildecviu,richard2017weakly,ectc,richard2018nnviterbi,isba,d3tw,CDFL}. Due to their cheap and fast acquisition, \emph{video transcripts} are a popular form of weak supervision \cite{hildecviu,richard2017weakly,richard2018nnviterbi,d3tw,CDFL}. In this case, only the order of occurring actions is given for training instead of frame-level labels. 

In this technical report, we take a closer look at the evaluation protocol for weakly supervised action segmentation using transcripts. In particular, we consider the recent approaches NNV \cite{richard2018nnviterbi}, ISBA \cite{isba}, CDFL \cite{CDFL} and MuCon \cite{mucon2021} for which source code is publicly available\footnote{\url{https://github.com/Zephyr-D/TCFPN-ISBA}\\ \url{https://github.com/alexanderrichard/NeuralNetwork-Viterbi}\\ \url{https://github.com/JunLi-Galios/CDFL}\\
\url{https://github.com/yassersouri/MuCon}}. 

Furthermore, we study the features used in weakly supervised action segmentation approaches. For the Breakfast dataset \cite{breakfast}, most approaches use IDT  features \cite{idt} whereas I3D features \cite{i3d} are extracted for the more recent CrossTask dataset \cite{crosstask}. 

From the experiments, we make the following observations:
\begin{itemize}
    \item The variance in the performance of weakly supervised approaches is high when running the training and testing multiple times with different random seeds.
    It is therefore necessary to report the average and standard deviation over multiple runs to evaluate the performance of an approach. 
    \item Unexpectedly, higher-level I3D features do not perform better than low-level IDT features. 
\end{itemize}

\section{Reproducibility Experiments}
\begin{table*}
    \centering
    \begin{tabular}{l|c|c|c|c}
        Model & \multicolumn{1}{p{1.5cm}|}{\centering MoF \\ Reported} &
                                \multicolumn{1}{p{1.75cm}|}{\centering MoF \\ Avg ($\pm$ Std)} &
                                \multicolumn{1}{p{1cm}|}{\centering MoF \\ Max} &
                                \multicolumn{1}{p{1cm}}{\centering MoF \\ Min}\\
        \Xhline{4\arrayrulewidth}
        ISBA \cite{isba}                                                 & 38.4     & 36.4 ($\pm$ 1.0)     & 37.6     & 35.1\\
        NNV \cite{richard2018nnviterbi}                 & 43.0     & 39.7 ($\pm$ 2.4)    & 43.5     & 37.5\\
        CDFL \cite{CDFL}                                                 & 50.2     & 48.1 ($\pm$ 2.5)     & 50.9     & 44.6\\
        MuCon \cite{mucon2021}                                                 & 48.5     & 48.5 ($\pm$ 1.7)     & 49.9     & 45.6\\
    \end{tabular}
    \vspace{1mm}
    \caption{
    Weakly supervised action segmentation performance on the Breakfast dataset. The mean over frame (MoF) is averaged over all dataset splits and $5$ different runs. `MoF Reported' denotes the reported MoF value from the corresponding paper.}
    \label{tab:results}
\end{table*}

For the experiments, we used the official source code for NNV \cite{richard2018nnviterbi}, ISBA \cite{isba}, and CDFL \cite{CDFL}. We run the training and testing on the Breakfast dataset without any modification to the respective source codes. As the parameters of the models published by NNV, ISBA, and CDFL were only provided for the Breakfast \cite{breakfast} dataset, we only perform experiments on this dataset. We perform 5 different iterations for each of the 4 splits of the Breakfast dataset and report the results in Table~\ref{tab:results}. We use the standard mean accuracy over frames (MoF) as the accuracy measure.

We observe that the average over 5 runs is a little bit lower than the reported results for all three methods, but the reported results are within the standard deviation. We also observe that the standard deviation is relatively high. For NNV \cite{richard2018nnviterbi} and CDFL \cite{CDFL}, it is about $2.5$. This shows that the methods for weakly supervised learning are more sensitive to the random model initialization and the random sampling of mini-batches compared to fully supervised approaches. It is therefore important to report the accuracy and standard deviation over multiple runs.  

\section{Effect of Feature Type}
The Breakfast dataset \cite{breakfast} is a standard benchmark for weakly supervised action segmentation and contains breakfast related preparation videos that were recorded in different kitchens. The more recent CrossTask dataset \cite{crosstask} is collected from YouTube videos and contains videos from different domains. While IDT features \cite{idt} are commonly used for the Breakfast dataset, \cite{crosstask} suggests to use I3D features \cite{i3d}. In general, it is expected that I3D features perform better since they are pre-trained on YouTube videos, but the Breakfast dataset differs from YouTube videos and contains videos with much more challenging light conditions. We, therefore, evaluated whether weakly supervised approaches benefit from I3D features on the Breakfast dataset.

\begin{table}[]
    \centering
    \begin{tabular}{c|c|c}
         Approach & Features & Average MoF \\
         \Xhline{4\arrayrulewidth}
         NNV  & IDT                  & 40.6\\
         NNV  & I3D                  & 11.4\\
         NNV  & PCA-I3D              & 23.2\\
         \hline
         CDFL & IDT                  & 48.9\\
         CDFL & I3D                  & 34.9\\
         CDFL & PCA-I3D              & 38.0\\
         \hline
         MuCon & IDT                  & 40.2\\
         MuCon & I3D                  & 48.3\\
         MuCon & PCA-I3D              & 47.7\\
    \end{tabular}
    \caption{Performance of different feature types. Average MoF is computed by running the training and testing of each method 5 times on split 1 of the Breakfast dataset.}
    \label{tab:i3d}
\end{table}

We extracted frame-wise I3D (RGB and Flow) features from the videos in the Breakfast dataset, obtaining a feature vector of size 2048 for each frame. We then re-run the experiments but only on split 1 of the Breakfast dataset.
% We use only the two best-performing methods NNV and CDFL.
The results are reported in Table~\ref{tab:i3d}.

We observe that without modification of the hyper-parameters NNV and CDFL cannot adapt to the new feature type and do not achieve a good performance.
We observe that MuCon performs poorly using the IDT features.
A major difference between the IDT features as they are used by NNV and CDFL compared to I3D is the feature dimensionality. We, therefore, reduced the dimensionality by PCA from 2048 to 64. We denote these features as PCA-I3D. Note that NNV and CDFL also reduce the dimensionality of IDT features to 64. The results in Table~\ref{tab:i3d} show that the dimensionality of the features has indeed an impact on the performance for both methods, but PCA-I3D is still worse than IDT.       

\begin{table}[]
    \centering
    \begin{tabular}{c|c|c}
         Approach & Feature Window Size & Average MoF \\
         \Xhline{4\arrayrulewidth}
         NNV  & 21                  & 23.2\\
         NNV  & 11                  & 24.6\\
         NNV  & 5                   & 26.3\\
         NNV  & 1                   & 33.9\\
         
    \end{tabular}
    \caption{Performance of NNV using PCA-I3D features with different feature window size. A smaller feature window size results in better accuracy. Average MoF is computed by running the training and testing 5 times on split 1 of the Breakfast dataset.}
    \label{tab:fws}
\end{table}

Another difference between IDT and I3D is the number of neighboring frames that are used to compute a feature vector for a frame. We, therefore, modified the temporal window of the GRU input for NNV. The results of our experiments with different feature window sizes are reported in Table~\ref{tab:fws}. They show that a smaller feature window size performs better when using PCA-I3D features. This indicates that current methods that perform well for weakly supervised action segmentation on Breakfast using IDT features cannot be directly applied to other features like I3D. Although the experiments do not fully reveal why this is the case, there are three possible explanations: a) I3D features do not work well on Breakfast in general, b) the existing methods need to be adapted to I3D, c) there is a need for different methods that benefit from I3D features.   

\section{Conclusion}
In this report, we have taken a closer look at two crucial aspects of recent weakly-supervised action segmentation approaches performance stability over multiple training runs and the impact of feature extraction. By performing experiments using the officially published source code of the published works, we find that the variance of the accuracy is relatively high. It is therefore necessary to report average and standard deviation over multiple runs to assess the performance of an approach for weakly supervised action segmentation. Furthermore, we observed that current methods do not benefit from `better' features. However, it remains an open research question why this is the case.   

{\small
\bibliographystyle{ieee_fullname}
\bibliography{egbib}
}

\end{document}